\documentclass[11pt]{article}

\usepackage[final]{acl}

\usepackage{times}
\usepackage{latexsym}

\usepackage[T1]{fontenc}

\usepackage[utf8]{inputenc}

\usepackage{microtype}

\usepackage{inconsolata}

\usepackage{graphicx}

\usepackage{booktabs} 
\usepackage{multirow} 
\usepackage{graphicx}
\usepackage{tcolorbox}
\usepackage{amsmath}
\usepackage{xspace}
\usepackage{makecell}
\usepackage{amssymb}
\usepackage{enumitem}

\newcommand{\framework}{\textsc{MLE-Ideator}\xspace}
\newcommand{\frameworkrl}{\textsc{MLE-Ideator-RL}\xspace}

\usepackage{xcolor}
\usepackage[table]{xcolor}
\definecolor{lightgray}{gray}{0.9} 
\definecolor{implementerFrame}{RGB}{60, 80, 100}
\definecolor{implementerBack}{RGB}{245, 247, 250}
\definecolor{ideatorFrame}{RGB}{40, 120, 60}
\definecolor{ideatorBack}{RGB}{240, 250, 240}
\tcbuselibrary{breakable}
\usepackage{listings}
\title{Learning to Ideate for Machine Learning Engineering Agents}

\author{
    Yunxiang Zhang\textsuperscript{1,2}\thanks{Work done during an internship at Amazon.} \quad 
    Kang Zhou\textsuperscript{1}\thanks{Corresponding authors.} \quad 
    Zhichao Xu\textsuperscript{1} \quad 
    Kiran Ramnath\textsuperscript{1} \\ 
    \quad \textbf{Yun Zhou}\textsuperscript{1} \quad 
    \textbf{Sangmin Woo}\textsuperscript{1} \quad 
    \textbf{Haibo Ding}\textsuperscript{1}\footnotemark[2] \quad 
    \textbf{Lin Lee Cheong}\textsuperscript{1} \\
    \textsuperscript{1}AWS AI Labs \quad \textsuperscript{2}University of Michigan \\
    \texttt{\{yxzyx,zhoukang,xzhichao,raxkiran,yunzzhou,}\\
    \texttt{sangminw,hbding,lcheong\}@amazon.com}
}

\begin{document}

\maketitle
\begin{abstract}

Existing machine learning engineering (MLE) agents  struggle to iteratively optimize their implemented algorithms for effectiveness. To address this, we introduce \textbf{\framework}, a dual-agent framework that separates ideation from implementation. In our system, an implementation agent can request strategic help from a dedicated \textsc{Ideator}. We show this approach is effective in two ways. First, in a training-free setup, our framework significantly outperforms implementation-only agent baselines on MLE-Bench. Second, we demonstrate that the \textsc{Ideator} can be trained with reinforcement learning (RL) to generate more effective ideas. With only 1K training samples from 10 MLE tasks, our RL-trained Qwen3-8B \textsc{Ideator} achieves an 11.5\% relative improvement  compared to its untrained counterpart and surpasses Claude Sonnet 3.5. 
These results highlights a promising path toward training strategic AI systems for scientific discovery.
\end{abstract}

\section{Introduction} 

\begin{figure}
    \centering 
    \includegraphics[width=0.7\linewidth]{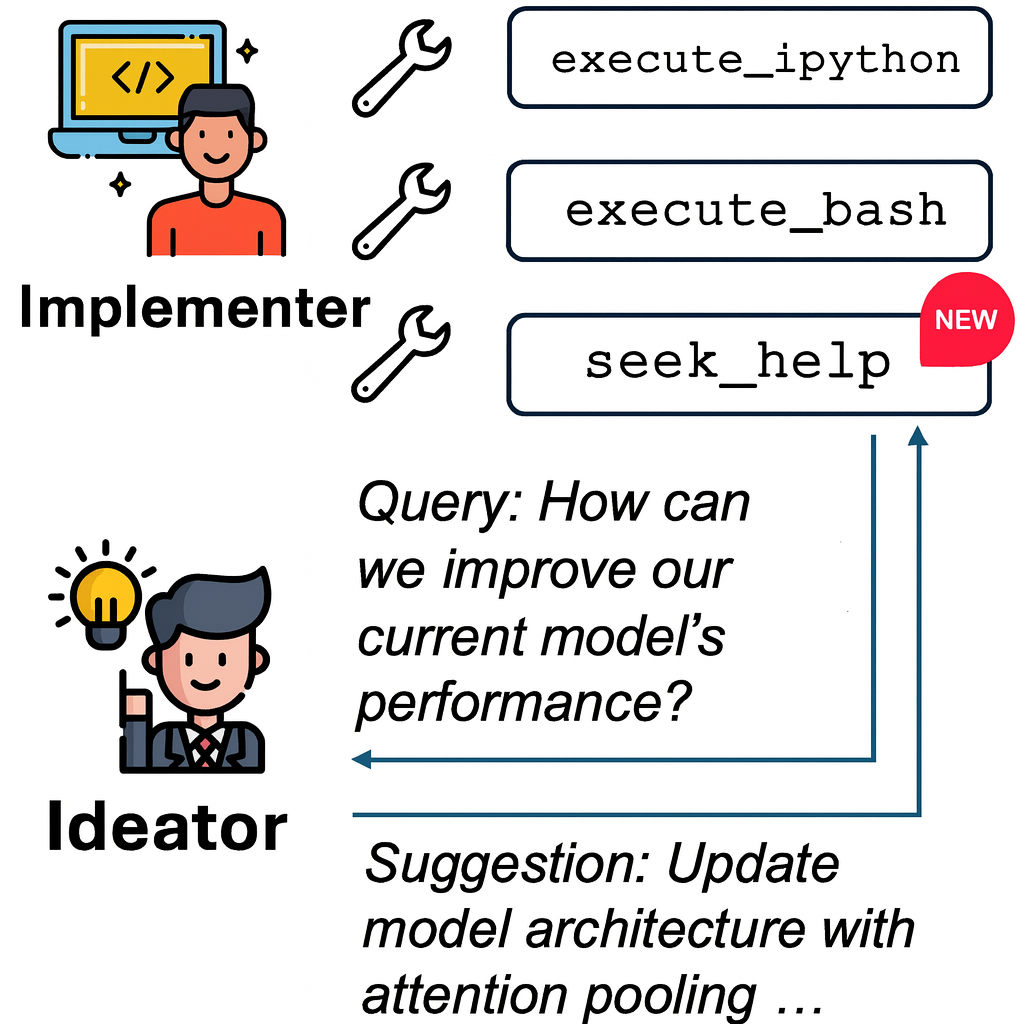}
\caption{
Overview of \textbf{\framework}. The \textsc{Implementer} uses a \texttt{<seek\_help>} action to solicit strategic guidance from a dedicated \textsc{Ideator}. This separation improves performance over single-agent baselines and enables the \textsc{Ideator} to be optimized via reinforcement learning with execution-based rewards.
}
    \label{fig:method} 
\end{figure}

Artificial intelligence (AI) agents capable of building machine learning models offer great potential to accelerate scientific discovery~\cite{Swanson2024.11.11.623004,DBLP:journals/corr/abs-2408-06292,DBLP:conf/iclr/ChanCJASMSLMPMW25}. While AI agents powered by large language models (LLMs) have made significant progress in relevant fields such as software development~\cite{opendevin,DBLP:conf/iclr/JimenezYWYPPN24,DBLP:conf/nips/YangJWLYNP24}, their typical designs are insufficient for autonomous machine learning engineering (MLE). They are often limited to implementing a single valid solution, ceasing exploration prematurely rather than continuously optimizing algorithms to enhance performance. In MLE tasks, a merely valid solution is insufficient, as the task demands strategic iteration—testing alternative models and hyperparameters, refining data and features to achieve state-of-the-art results~\citep{DBLP:conf/iclr/ChanCJASMSLMPMW25,DBLP:journals/corr/abs-2505-07782,DBLP:journals/corr/abs-2504-09702}.

While recent work has sought to develop agents capable of strategic iteration, typically  by training small language models (SLMs) as MLE agents via reinforcement learning (RL)~\cite{DBLP:journals/corr/abs-2505-23723,yang2025reinforcement},
these approaches typically rely on a single-agent paradigm that conflates strategic reasoning (idea proposal) with coding proficiency (implementation). This conflation imposes a critical trade-off: small models lack the coding capacity to implement complex solutions, yet training them to master both skills requires inefficient multi-turn rollouts with sparse rewards. To address this structural limitation, we ask: \textit{Can decoupling ideation from implementation enable small models to efficiently learn strategies that guide stronger models?}

We answer this question by designing a novel dual-agent framework \textbf{\framework} (Figure~\ref{fig:method}) that explicitly separates ideation from implementation. When the Implementer's progress plateaus, it can invoke a new \texttt{<seek\_help>} action to solicit high-level strategies from the \textsc{Ideator}. This design facilitates  \textit{superalignment}~\cite{DBLP:conf/icml/BurnsIKBGACEJLS24}, where a lightweight \textsc{Ideator} model serves as a strategic guider for a more capable Implementer, directing algorithm refinements without needing to retrain the Implementer itself. A well-trained \textsc{Ideator} can thus be reused as a plug-in to augment the performance of various implementer models.

 We first demonstrate on the MLE-Bench~\cite{DBLP:conf/iclr/ChanCJASMSLMPMW25} that training-free ideation improves performance over implementation-only agent baselines. To further enhance ideation capability, we introduce an RL pipeline to train the \textsc{Ideator}. Each proposed idea is executed by a frozen Implementer in a \textit{single} step and rewarded if it improves performance. The \textsc{Ideator} is then updated using the GRPO~\cite{DBLP:journals/corr/abs-2402-03300} algorithm to maximize this execution-based reward, thereby efficiently avoiding costly multi-step rollouts. Our experiments show that the Qwen3-8B model, after RL training on just 10 MLE tasks, not only outperforms its untrained counterpart but also surpasses the powerful Claude Sonnet 3.5 as an \textsc{Ideator}. These results highlight a promising path toward teaching LLM agents to generate high-impact ideas for scientific discovery.

Our contributions are threefold: 
\begin{itemize}
    \item We introduce \framework, a novel dual-agent framework that separates high-level ideation from low-level implementation for MLE tasks.
    \item We design an efficient RL pipeline to train an \textsc{Ideator} with execution-based rewards, minimizing reliance on costly multi-step rollouts.
    \item We demonstrate strong performance gains on MLE-Bench, showing that an RL-trained small Ideator can outperform more powerful LLMs in guiding the Implementer.
\end{itemize}

\section{Related Work}
\label{app:related_work}

\paragraph{Single-Agent Frameworks for MLE.} Coding agents such as CodeAct~\cite{opendevin}, MLAB~\cite{Huang2023MLAgentBenchEL}, and RepoMaster~\cite{DBLP:journals/corr/abs-2505-21577} execute MLE tasks using basic tools like Python and Bash but lack dedicated ideation, often stopping at the first valid solution. Parallel exploration frameworks like AIDE~\cite{Jiang2025AIDEAE}, AIRA~\cite{DBLP:journals/corr/abs-2507-02554}, and ML-Master~\cite{DBLP:journals/corr/abs-2506-16499} extend this via tree search across multiple solution branches. However, these methods rely on expensive code-space search and static prompting, limiting exploration to implementation rather than high-level ideation. In contrast, we decouple ideation from coding, enabling efficient idea-space exploration without costly tree-search orchestration.

\paragraph{Multi-Agent Frameworks for MLE.}
Multi-agent frameworks like MLZero~\cite{DBLP:journals/corr/abs-2505-13941}, InternAgent~\cite{Zhang2025NovelSeekWA}, Agent Laboratory~\cite{DBLP:journals/corr/abs-2501-04227}, and R\&D-Agent~\cite{DBLP:journals/corr/abs-2505-14738} decompose tasks using specialized roles (e.g., planner, experimenter). Yet, their coordination relies on complex, rigid, and prompt-driven workflows. Our work differs in two ways. First, we employ a dynamic workflow where the Implementer solicits on-demand strategic guidance via a \texttt{<seek\_help>} action, ensuring collaboration is context-driven rather than predefined. Second, unlike prompt-only systems, we apply reinforcement learning directly to the ideation model, optimizing it to generate effective algorithmic suggestions based on performance outcomes.

\paragraph{RL-Trained MLE Agents.} Reinforcement learning has recently been applied to MLE agents~\cite{DBLP:journals/corr/abs-2505-23723,yang2025reinforcement}, but these works focus on training single-agent implementers for code generation, neglecting the strategic challenge of ideation. By contrast, we train the \textsc{Ideator} itself with RL to directly optimize idea quality via performance feedback. This lightweight mechanism enables the \textsc{Ideator} to learn high-level refinements. 

\section{Methods}
\subsection{\framework}
To decouple ideation from implementation in machine learning engineering tasks, we propose a dual-agent framework in Figure~\ref{fig:method} that consists of two primary components: a Implementer agent responsible for implementation and an \textsc{Ideator} agent for providing high-level suggestions.
We instantiate the Implementer with the CodeAct agent framework~\cite{opendevin}, which provides an action space for executing Python and Bash commands (\texttt{<execute\_ipython>} and \texttt{<execute\_bash>}). We introduce a new action, \texttt{<seek\_help>}, which the Implementer can invoke as needed when it  experiences plateaued performance, or identifies a potential optimization opportunity. The query is formatted as follows:

\begin{tcolorbox}[
    colback=black!5!white,         
    colframe=black!75!white,       
    title=Implementer Agent Query via <seek\_help>, 
    fonttitle=\bfseries\small,           
    boxrule=1pt,                   
    arc=2mm,                       
    left=2mm
]
\scriptsize
\begin{verbatim}
<seek_help>
PROBLEM_STATEMENT:
<one sentence describing the current blocking issue>

ATTEMPTS_SO_FAR:
<a short bullet list of what you already tried>

GOAL:
<one sentence on what success looks like for this step>
</seek_help>
\end{verbatim}


\end{tcolorbox}

The \textsc{Ideator} receives this query along with the full trajectory of the Implementer's work (including code, logs,  performance) and provides a structured suggestion:

\begin{tcolorbox}[
    colback=black!5!white,         
    colframe=black!75!white,       
    title=\textsc{Ideator} Agent Response for <seek\_help>, 
    fonttitle=\bfseries\small,           
    boxrule=1pt,                   
    arc=2mm,                       
    left=2mm
]
\scriptsize
\begin{verbatim}
ANALYSIS_ON_CURRENT_PROGRESS:
<Briefly state whether to keep refining the present 
approach or revert to a prior solution and pursue a
new path.>

ACTION:
<One imperative command or code block the agent must 
execute next.>

RATIONALE:
<Concise justification for why this action is optimal
right now.>
\end{verbatim}


\end{tcolorbox}

This structure makes the suggestions easily comprehendable and actionable for the Implementer.
The Implementer receives the \textsc{Ideator}'s output as an observation and integrates the suggested action into its workflow in the subsequent step. This protocol enables targeted, context-aware collaboration.

\subsection{\frameworkrl}

To enable the \textsc{Ideator} to learn from experience, we introduce a reinforcement learning framework to train the \textsc{Ideator} to propose more effective ideas.

\paragraph{State and Action Space.}
At each timestep $t$ where the Implementer issues a \texttt{<seek\_help>} action, the \textsc{Ideator} receives a comprehensive snapshot of the Implementer agent's progress as the current state representation $s_t = (\mathcal{D}, \tau_{1:t}, p_t, C_t)$, consisting of the task description $\mathcal{D}$, the current trajectory $\tau_{1:t} = (a_1, o_1, ..., a_t)$, which includes all previous actions and observations, the current ML solution's performance score $p_t$, and the full ML solution code $C_t$.
The \textsc{Ideator}'s action is to generate a natural language suggestion $\alpha$ with a structured format: $\alpha = (\text{ANALYSIS}, \text{ACTION}, \text{RATIONALE})$. 

\paragraph{Reward Function.} To directly optimize for effective ideas, we define a discrete, three-level reward function $R_t(\alpha)$ that evaluates the outcome of applying an idea $\alpha$:
\begin{equation}\label{eq:reward}
R_t(\alpha) = 
\begin{cases}
+1 & \text{if } p_{t+1}(\alpha) > p_t \\
0  & \text{if } p_{t+1}(\alpha) \le p_t \text{ or execution fails} \\
-1 & \text{if format errors,}
\end{cases}
\end{equation}
where $p_t$ is the solution performance before the idea, and $p_{t+1}(\alpha)$ is the new performance after the Implementer executes  revised code based on the idea $\alpha$ in a \textit{single} step. 
The reward function incentivizes the \textsc{Ideator} to propose ideas that improve performance and penalizes those that are ineffective or invalid.

\paragraph{RL Training Pipeline.} 
To avoid regenerating agent trajectories online at each training step, we first run \framework offline to collect a pool of states whose trajectories end with the \texttt{<seek\_help>} action. These states are then used to construct prompts for the \textsc{Ideator} (see Appendix~\ref{app:prompt_ideator}). We employ GRPO~\cite{DBLP:journals/corr/abs-2402-03300} as our RL algorithm for its efficiency and stability. The training objective is to maximize the expected reward for the ideas generated by the \textsc{Ideator} (Appendix~\ref{app:rl_obj}). Each training step begins with an \textbf{ideation (rollout) phase}, where we sample $G$ candidate ideas $\{\alpha_1, ..., \alpha_G\}$ from the \textsc{Ideator}'s current policy. Next, during the \textbf{execution (reward calculation) phase}, a frozen Implementer agent attempts to implement each valid idea in a single step.\footnote{We restrict execution to a single step for simplicity and efficiency, though more turns could reduce execution failure through agent self-debugging.} If execution succeeds, we evaluate the new ML solution to compute a performance-based reward defined in Equation~\ref{eq:reward}. Finally, in the \textbf{back-propagation phase}, we update the \textsc{Ideator}'s policy parameters using the GRPO loss function (Appendix~\ref{app:grpo_loss}). This formulation enables direct optimization of idea effectiveness through performance-based feedback.

\section{Experiments}

\subsection{Experimental Setup}
\paragraph{Dataset and Baselines.}
Our experiments are conducted on the \textbf{MLE-Bench} dataset~\cite{DBLP:conf/iclr/ChanCJASMSLMPMW25}. For our training tasks, we select 10 diverse Kaggle tasks (Appendix~\ref{app:training_task_selection}) that span multiple data modalities and whose solutions can typically be executed within 20 minutes, a criterion that enables faster reward calculation.
To collect training data from these 10 tasks, we run the \framework framework 90 times for each using Claude Sonnet 3.5 as the backbone model and randomly sample 1000 states to serve as training prompts (100 per task) and 100 states for validation (10 per task).\footnote{Each run may trigger multiple \texttt{<seek\_help>} calls, producing over 1,000 states ending with that action. We sample the required training and validation sets from this pool.}
Our evaluation is performed on a held-out set of 51 tasks (21 low-complexity and 30 medium-complexity), excluding the 15 high-complexity tasks where agents consistently struggle to produce a valid submission within the time limit. We test several configurations using Qwen3-8B~\cite{DBLP:journals/corr/abs-2505-09388} and Claude Sonnet 3.5 as the backbone LLMs for the \textsc{Ideator} and Implementer roles. We compare three configurations: a \textbf{CodeAct} baseline where the Implementer agent solves the task alone; an \textbf{\framework (Prompting)} setup where the Implementer is paired with a prompted \textsc{Ideator} and can use the \texttt{<seek\_help>} action; and our \textbf{\frameworkrl (RL-Trained)} approach, which replaces the prompted Qwen3-8B \textsc{Ideator} with our RL-trained version. During evaluation, all agents are allotted a maximum of 50 steps and a one-hour runtime per task. We show the hyperparameters of RL training in Appendix~\ref{app:training_hyperparam}.

\begin{table}
    \centering
    \small
    \setlength{\tabcolsep}{3pt}
    \begin{tabular}{lccc} 
        \toprule
        \textbf{System} & \textbf{\textsc{Ideator}} & \textbf{Avg@3} & \textbf{Best@3} \\
        \midrule
        \rowcolor{lightgray}
        \multicolumn{4}{c}{\textit{\textbf{Implementation-only Agent Baselines}}}\\\midrule
        CodeAct + GPT-4o & - & 47.9 & 51.7 \\
        AIDE + GPT-4o & - & 49.6 & 50.7 \\
        AIDE + Sonnet 3.5 & - & 50.7 & 53.8 \\
        \midrule
        
        \rowcolor{lightgray}
        \multicolumn{4}{c}{\textit{\textbf{Implementer = Claude Sonnet 3.5}}} \\\midrule
        \makecell[l]{CodeAct} & - & 50.6 & 52.8 \\\cmidrule{1-2}
        \multirow{2}{*}{\makecell[l]{\framework}} & Sonnet 3.5 & \textbf{58.5} & \underline{60.9} \\
       & Qwen3-8B          & 53.2 & 56.6 \\\cmidrule{1-2}
        \makecell[l]{\frameworkrl} & Qwen3-8B-RL & \underline{58.4} & \textbf{63.1} \\
        \midrule
        \rowcolor{lightgray}
        \multicolumn{4}{c}{\textit{\textbf{Implementer = Qwen3-8B}}} \\\midrule
        \makecell[l]{CodeAct} & - & 25.4 & 25.9 \\\cmidrule{1-2}
        \multirow{2}{*}{\makecell[l]{\framework}} & Sonnet 3.5 & \underline{28.0} & \underline{28.3} \\
              & Qwen3-8B          & 25.2 & 25.6 \\\cmidrule{1-2}
         \makecell[l]{\frameworkrl} & Qwen3-8B-RL & \textbf{29.8} & \textbf{30.1} \\
        \bottomrule
    \end{tabular}
        \caption{
        Main results on 51 held-out MLE-Bench tasks.
        }
    \label{tab:ideator_performance}
\end{table}

\paragraph{Evaluation Metric.}

We evaluate agent performance using \textbf{Avg@3} and \textbf{Best@3} normalized scores~\cite{jiang2025bootstrapping}, averaged across all tasks.
For each task, an agent is run three independent times.
Since the native evaluation metrics (e.g., accuracy, cross-entropy loss) vary in scale from one Kaggle task to another, we normalize the raw score of each run against human performance on the Kaggle leaderboard to a common 0-100 scale:
$\max(0, 100 \times \frac{\text{agent\_score} - \text{worst\_human\_score}}{\text{best\_human\_score} - \text{worst\_human\_score}})$.
The Best@3 is the maximum of these three normalized scores, while the Avg@3 is their average, excluding any runs that failed to produce a valid submission. An agent receives a score of zero for a task only if all three runs fail to produce any valid solutions.

\subsection{Results and Analysis}
\paragraph{Prompted and Reinforced Ideation Boost Performance.}
Table~\ref{tab:ideator_performance} shows that prompted ideation consistently boosts performance over implementation-only agent baselines. With Claude Sonnet 3.5 as the Implementer, pairing it with a prompted \textsc{Ideator} yields strong improvements: Avg@3 rises from 50.6 (CodeAct baseline) to 58.5 with a Sonnet \textsc{Ideator} and 53.2 with a Qwen3-8B \textsc{Ideator}. Reinforcement learning further enhances effectiveness, as the Qwen3-8B-RL \textsc{Ideator} achieves a Best@3 of 63.1, surpassing even the powerful Sonnet \textsc{Ideator}. 
Even with the weaker Qwen3-8B Implementer, ideation proves valuable: pairing with Sonnet 3.5 lifts performance from 25.4 to 28.0, and the RL-trained Qwen3-8B \textsc{Ideator} achieves a stronger result at 29.8.
These results show that \textit{prompted ideation consistently improves implementers, and RL enables smaller \textsc{Ideators} to surpass stronger prompted ones.}

\begin{table}
\centering
\small
\begin{tabular}{ccc}
\toprule
\textbf{\textsc{Ideator}} & \textbf{Avg@3} & \textbf{Best@3} \\
\midrule
\rowcolor{lightgray}
\multicolumn{3}{c}{\textit{\textbf{Implementer = Claude Sonnet 3.5}}} \\\midrule
-  & 69.7 & 72.0 \\
\textsc{Null Idea} & 68.7 & 72.8 \\
\textsc{Vague Idea} & 75.0 & 76.3 \\
Claude Sonnet 3.5 & \textbf{80.1} & \textbf{83.8} \\
\bottomrule
\end{tabular}
\caption{
Impact of idea quality on \framework performance under 22 low-complexity MLE-Bench tasks.
\textsc{Null Idea} and \textsc{Vague Idea} denote non-LLM ablations where the \textsc{Ideator} is replaced with a fixed template response.
\textsc{Null Idea} always returns ``\textit{I have no suggestions for improving the solution. Please proceed using your best judgment.}'' and \textsc{Vague Idea} always returns ``\textit{Keep improving the performance of your solution.}''
}
\label{tab:ablation_study}
\end{table}

\paragraph{Idea Quality Matters.}
Is the performance boost driven by the quality of ideas, or simply by adding the \texttt{<seek\_help>} action and receiving some form of feedback, even if it is uninformative?
Table~\ref{tab:ablation_study} ablates this factor by replacing the LLM-based \textsc{Ideator} with fixed template outputs.
Using a \textsc{Null Idea} yields performance comparable to having no \textsc{Ideator}, since the Implementer receives no meaningful guidance beyond its own trajectory.
A \textsc{Vague Idea}, which always provides a generic encouragement, produces a slight improvement, likely because it prompts the Implementer to continue refining rather than stopping prematurely.
However, this effect is small compared to the substantial gains achieved with specific, contextual ideas from the Claude Sonnet 3.5 \textsc{Ideator}.
These results confirm that \textit{high-quality, actionable guidance, rather than reflection or additional interaction, drives the performance improvement of \framework.}

\begin{table}
\centering
\small
\setlength{\tabcolsep}{2pt}
\begin{tabular}{lccc}
\toprule
\textbf{Idea Type} & \textbf{Qwen3-8B} & \textbf{Qwen3-8B-RL} & $\Delta$ \\
\midrule
Data Preparation     & 13.4 & 16.5 & \cellcolor{green!10} +3.1 \\
Feature Engineering  & 13.4 & 20.9 & \cellcolor{green!20} +7.5 \\
Model Architecture   & 28.5 & 25.3 & \cellcolor{red!15} -3.2 \\
Model Training       & 32.4 & 27.5 & \cellcolor{red!20} -4.9 \\
Hyperparameter Tuning& 7.3  & 6.6  & \cellcolor{red!5} -0.7 \\
Others               & 5.0  & 3.3  & \cellcolor{red!10} -1.7 \\
\bottomrule
\end{tabular}
\caption{Comparison of idea type distributions between Qwen3-8B and Qwen3-8B-RL. Values indicate the proportion (\%) of ideas in each category, with $\Delta$ showing changes after RL training (cells shaded \colorbox{green!25}{green for increases} and \colorbox{red!25}{red for decreases}). 
}
\label{tab:idea_shift}
\end{table}

\paragraph{RL Aligns Idea Generation with Empirically Effective Strategies.}
Not all ideas contribute equally to performance, so we analyze which types are most effective. We define an idea as \textit{effective} if it improves the ML solution’s performance. To this end, we prompt a strong LLM (Claude Sonnet 4) to classify each idea into six categories: Data Preparation, Feature Engineering, Model Architecture, Model Training, Hyperparameter Tuning, and Others (full prompt in Appendix~\ref{app:prompt_classification}).
As shown in Figure~\ref{fig:idea_effectiveness} (Appendix~\ref{app:idea_type_analysis}), \textbf{Feature Engineering} and \textbf{Data Preparation} ideas are typically more effective, whereas \textbf{Model Training} and \textbf{Hyperparameter Tuning} are less reliable.
Building on this observation, Table~\ref{tab:idea_shift} shows that \textit{RL training shifts the idea distribution toward empirically effective categories} and reduces the frequency of less effective ones, thereby improving overall idea quality.
We show concrete examples of the \textsc{Ideator}'s successful and unsuccessful suggestions in Appendix~\ref{app:case_study}.

\begin{figure}
    \centering 
    \includegraphics[width=\linewidth]{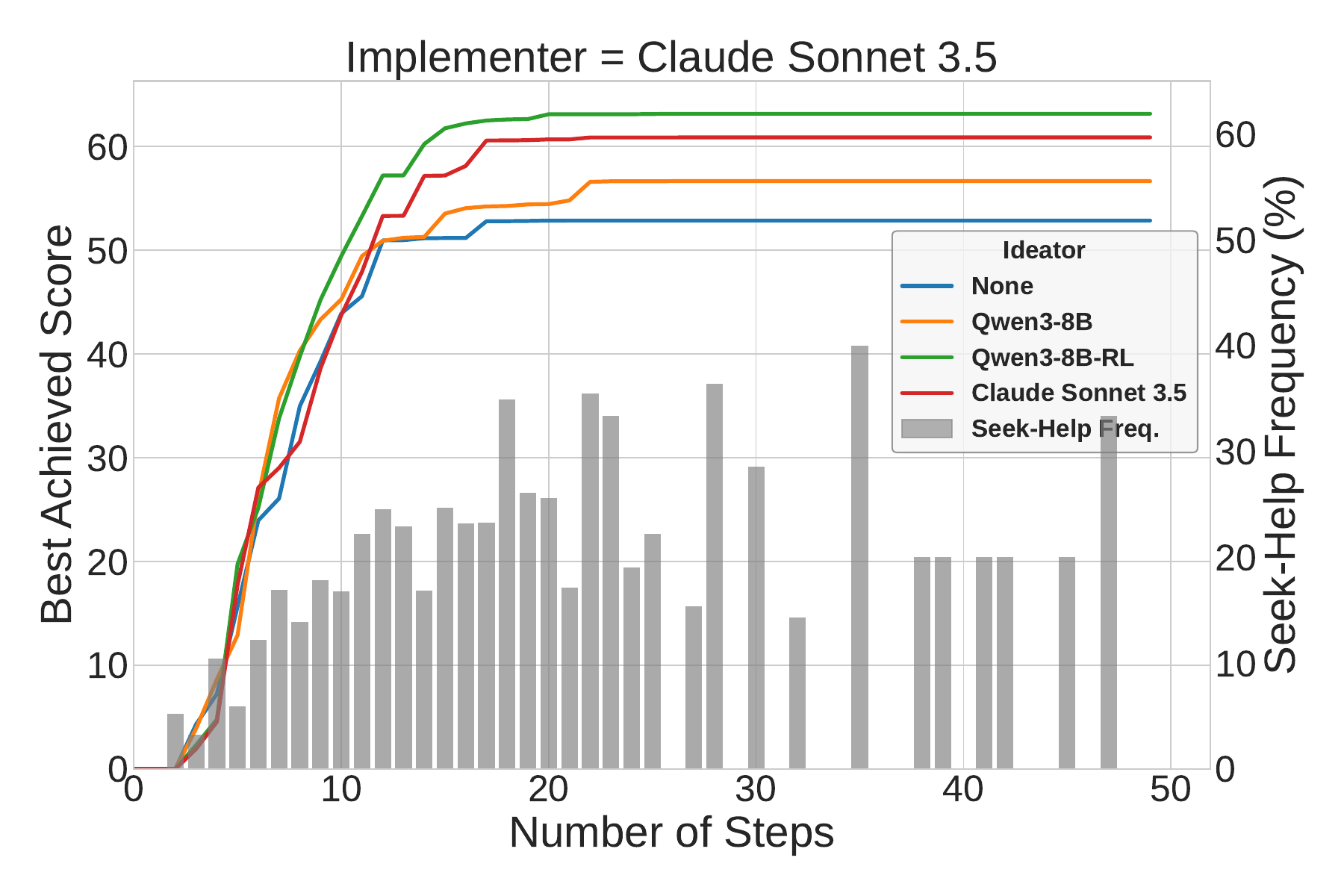}
    \caption{For each \textsc{Ideator} model, we show the agent's best achieved normalized score so far in a trajectory averaged over all tasks w.r.t the number of steps in trajectory. We also plot the frequency of \texttt{<seek\_help>} actions at each steps, aggregated over all \textsc{Ideators}.} 
    \label{fig:agent_step_scores} 
\end{figure}

\paragraph{Ideation Drives Mid-Trajectory Refinement.}
Figure \ref{fig:agent_step_scores} correlates agent performance with \texttt{<seek\_help>} behavior. Initially (<12 steps), the Implementer focuses on establishing a baseline with minimal help-seeking. Once an initial submission is made, \texttt{<seek\_help>} frequency spikes. This \textsc{Ideator}-guided exploration drives immediate performance gains, allowing the system to escape local optima where single-agent baselines stagnate. Performance generally plateaus around step 20 due to diminishing returns on refinements and the one-hour execution limit. Future integration with inference-time scaling~\cite{DBLP:journals/corr/abs-2408-03314,DBLP:journals/corr/abs-2506-12928,DBLP:journals/corr/abs-2506-07976} could unlock further gains via Best-of-N idea sampling.

\section{Conclusion}

We introduce \framework, a dual-agent framework that decouples strategic ideation from implementation for machine learning engineering tasks. Our experiments on MLE-Bench demonstrate that this approach significantly outperforms implementation-only baselines. Furthermore, by training the Ideator with reinforcement learning using execution-based rewards, our Qwen3-8B model achieves an 11.5\% relative improvement over its prompted counterpart and surpasses the proprietary Claude Sonnet 3.5, successfully aligning ideation with high-impact strategies. Our work provides a promising training recipe for LLM agents that can  automate AI research.

\section*{Limitations}

\paragraph{Additional Inference Cost.}
The \framework framework incurs a higher inference cost than the implementation-only agent baseline, i.e., CodeAct~\cite{opendevin}. When using Claude Sonnet 3.5 for both the Implementer and \textsc{Ideator} roles, a single run of our framework costs 1.4 USD on average, compared to 0.9 USD for the baseline. This increased cost stems from two primary factors:
\begin{itemize}
    \item \textbf{Additional Steps:} The dual-agent system averages more steps per run (17.4 vs. 15.0) due to the introduction of the \texttt{<seek\_help>} action required to query the \textsc{Ideator}.
    \item \textbf{Expanded Context:} To provide the \textsc{Ideator} with sufficient history, its prompt includes the entire agent trajectory (truncated to 32,000 tokens). This significantly increases the number of input tokens per call.
\end{itemize}
We expect that this cost can be mitigated. Future work could focus on developing summarization or pruning techniques~\cite{DBLP:journals/corr/abs-2505-21577,DBLP:journals/corr/abs-2507-13334} to reduce the context length of the trajectory without sacrificing the quality of the generated ideas.

Additionally, the inference latency introduced by ideation is minor relative to the end-to-end ML pipeline, as each \texttt{<seek\_help>} action is just a lightweight LLM call compared to the far more time-consuming code execution, model training, and evaluation steps of each ML task.

\paragraph{Resource-Intensive RL Training.}
The reinforcement learning process is resource-intensive, as calculating the reward for each proposed idea requires executing a full ML solution, which often involves training a neural network on a GPU.
In our training setup, the reward calculation is distributed across 16 nodes, each with 8 A10G (24GB) GPUs, where each GPU is dedicated to running the solution for a single candidate idea (Appendix~\ref{app:training_hyperparam}).
Our use of 128 A10 GPUs was a choice to accelerate reward computation with parallel ML solution execution, but not a strict requirement. The same procedure can be run with fewer GPUs by executing candidate solutions sequentially, trading off training time for compute resources.

A primary bottleneck to scalability lies in the requirement of a full execution to evaluate each generated idea. To mitigate this limitation, future work could focus on developing a proxy reward model. Such a model, by predicting an idea's effectiveness without costly execution~\cite{park2025lookleapestimatingllm,DBLP:conf/naacl/AnugrahaWLIL25,DBLP:journals/corr/abs-2506-00794}, would significantly enhance the scalability of the training process.

\bibliography{custom}

@article{DBLP:journals/corr/abs-2504-09702,
  author       = {Yunxiang Zhang and
                  Muhammad Khalifa and
                  Shitanshu Bhushan and
                  Grant D. Murphy and
                  Lajanugen Logeswaran and
                  Jaekyeom Kim and
                  Moontae Lee and
                  Honglak Lee and
                  Lu Wang},
  title        = {{MLRC-B}ench: Can Language Agents Solve Machine Learning Research Challenges?},
  journal      = {CoRR},
  volume       = {abs/2504.09702},
  year         = {2025},
  url          = {https://doi.org/10.48550/arXiv.2504.09702},
  doi          = {10.48550/ARXIV.2504.09702},
  eprinttype    = {arXiv},
  eprint       = {2504.09702},
  bibsource    = {dblp computer science bibliography, https://dblp.org}
}

@article{DBLP:journals/corr/abs-2506-00794,
  author       = {Jiaxin Wen and
                  Chenglei Si and
                  Yueh{-}han Chen and
                  He He and
                  Shi Feng},
  title        = {Predicting Empirical {AI} Research Outcomes with Language Models},
  journal      = {CoRR},
  volume       = {abs/2506.00794},
  year         = {2025},
  url          = {https://doi.org/10.48550/arXiv.2506.00794},
  doi          = {10.48550/ARXIV.2506.00794},
  eprinttype    = {arXiv},
  eprint       = {2506.00794},
  bibsource    = {dblp computer science bibliography, https://dblp.org}
}

@article{DBLP:journals/corr/abs-2507-13334,
  author       = {Lingrui Mei and
                  Jiayu Yao and
                  Yuyao Ge and
                  Yiwei Wang and
                  Baolong Bi and
                  Yujun Cai and
                  Jiazhi Liu and
                  Mingyu Li and
                  Zhong{-}Zhi Li and
                  Duzhen Zhang and
                  Chenlin Zhou and
                  Jiayi Mao and
                  Tianze Xia and
                  Jiafeng Guo and
                  Shenghua Liu},
  title        = {A Survey of Context Engineering for Large Language Models},
  journal      = {CoRR},
  volume       = {abs/2507.13334},
  year         = {2025},
  url          = {https://doi.org/10.48550/arXiv.2507.13334},
  doi          = {10.48550/ARXIV.2507.13334},
  eprinttype    = {arXiv},
  eprint       = {2507.13334},
  bibsource    = {dblp computer science bibliography, https://dblp.org}
}

@article{yang2025reinforcement,
  title={Reinforcement Learning for Machine Learning Engineering Agents},
  author={Yang, Sherry and He-Yueya, Joy and Liang, Percy},
  journal={arXiv preprint arXiv:2509.01684},
  year={2025}
}

@article{DBLP:journals/corr/abs-2505-21577,
  author       = {Huacan Wang and
                  Ziyi Ni and
                  Shuo Zhang and
                  Shuo Lu and
                  Sen Hu and
                  Ziyang He and
                  Chen Hu and
                  Jiaye Lin and
                  Yifu Guo and
                  Yuntao Du and
                  Pin Lyu},
  title        = {RepoMaster: Autonomous Exploration and Understanding of GitHub Repositories
                  for Complex Task Solving},
  journal      = {CoRR},
  volume       = {abs/2505.21577},
  year         = {2025},
  url          = {https://doi.org/10.48550/arXiv.2505.21577},
  doi          = {10.48550/ARXIV.2505.21577},
  eprinttype    = {arXiv},
  eprint       = {2505.21577},
  bibsource    = {dblp computer science bibliography, https://dblp.org}
}

@article{jiang2025bootstrapping,
  title={Bootstrapping Task Spaces for Self-Improvement},
  author={Jiang, Minqi and Lupu, Andrei and Bachrach, Yoram},
  journal={arXiv preprint arXiv:2509.04575},
  year={2025}
}

@article{DBLP:journals/corr/abs-2505-14738,
  author       = {Xu Yang and
                  Xiao Yang and
                  Shikai Fang and
                  Bowen Xian and
                  Yuante Li and
                  Jian Wang and
                  Minrui Xu and
                  Haoran Pan and
                  Xinpeng Hong and
                  Weiqing Liu and
                  Yelong Shen and
                  Weizhu Chen and
                  Jiang Bian},
  title        = {R{\&}D-Agent: Automating Data-Driven {AI} Solution Building Through
                  LLM-Powered Automated Research, Development, and Evolution},
  journal      = {CoRR},
  volume       = {abs/2505.14738},
  year         = {2025},
  url          = {https://doi.org/10.48550/arXiv.2505.14738},
  doi          = {10.48550/ARXIV.2505.14738},
  eprinttype    = {arXiv},
  eprint       = {2505.14738},
  bibsource    = {dblp computer science bibliography, https://dblp.org}
}

@article{DBLP:journals/corr/abs-2501-04227,
  author       = {Samuel Schmidgall and
                  Yusheng Su and
                  Ze Wang and
                  Ximeng Sun and
                  Jialian Wu and
                  Xiaodong Yu and
                  Jiang Liu and
                  Zicheng Liu and
                  Emad Barsoum},
  title        = {Agent Laboratory: Using {LLM} Agents as Research Assistants},
  journal      = {CoRR},
  volume       = {abs/2501.04227},
  year         = {2025},
  url          = {https://doi.org/10.48550/arXiv.2501.04227},
  doi          = {10.48550/ARXIV.2501.04227},
  eprinttype    = {arXiv},
  eprint       = {2501.04227},
  bibsource    = {dblp computer science bibliography, https://dblp.org}
}

@article{Zhang2025NovelSeekWA,
  title={NovelSeek: When Agent Becomes the Scientist - Building Closed-Loop System from Hypothesis to Verification},
  author={Bo Zhang and Shi Feng and Xiangchao Yan and Jiakang Yuan and Zhiyin Yu and Xiaohan He and Songtao Huang and Shaowei Hou and Zheng Nie and Zhilong Wang and Jinyao Liu and Runmin Ma and Tianshuo Peng and Peng Ye and Dongzhan Zhou and Shufei Zhang and Xiaosong Wang and Yilan Zhang and Meng Li and Zhongying Tu and Xiangyu Yue and Wangli Ouyang and Bowen Zhou and Lei Bai},
  journal={ArXiv},
  year={2025},
  volume={abs/2505.16938},
  url={https://api.semanticscholar.org/CorpusID:278788499}
}

@article{DBLP:journals/corr/abs-2506-16499,
  author       = {Zexi Liu and
                  Yuzhu Cai and
                  Xinyu Zhu and
                  Yujie Zheng and
                  Runkun Chen and
                  Ying Wen and
                  Yanfeng Wang and
                  Weinan E and
                  Siheng Chen},
  title        = {ML-Master: Towards AI-for-AI via Integration of Exploration and Reasoning},
  journal      = {CoRR},
  volume       = {abs/2506.16499},
  year         = {2025},
  url          = {https://doi.org/10.48550/arXiv.2506.16499},
  doi          = {10.48550/ARXIV.2506.16499},
  eprinttype    = {arXiv},
  eprint       = {2506.16499},
  bibsource    = {dblp computer science bibliography, https://dblp.org}
}

@article{DBLP:journals/corr/abs-2505-13941,
  author       = {Haoyang Fang and
                  Boran Han and
                  Nick Erickson and
                  Xiyuan Zhang and
                  Su Zhou and
                  Anirudh Dagar and
                  Jiani Zhang and
                  Ali Caner T{\"{u}}rkmen and
                  Cuixiong Hu and
                  Huzefa Rangwala and
                  Ying Nian Wu and
                  Yuyang Wang and
                  George Karypis},
  title        = {MLZero: {A} Multi-Agent System for End-to-end Machine Learning Automation},
  journal      = {CoRR},
  volume       = {abs/2505.13941},
  year         = {2025},
  url          = {https://doi.org/10.48550/arXiv.2505.13941},
  doi          = {10.48550/ARXIV.2505.13941},
  eprinttype    = {arXiv},
  eprint       = {2505.13941},
  bibsource    = {dblp computer science bibliography, https://dblp.org}
}

@article{DBLP:journals/corr/abs-2507-02554,
  author       = {Edan Toledo and
                  Karen Hambardzumyan and
                  Martin Josifoski and
                  Rishi Hazra and
                  Nicolas Mario Baldwin and
                  Alexis Audran{-}Reiss and
                  Michael Kuchnik and
                  Despoina Magka and
                  Minqi Jiang and
                  Alisia Maria Lupidi and
                  Andrei Lupu and
                  Roberta Raileanu and
                  Kelvin Niu and
                  Tatiana Shavrina and
                  Jean{-}Christophe Gagnon{-}Audet and
                  Michael Shvartsman and
                  Shagun Sodhani and
                  Alexander H. Miller and
                  Abhishek Charnalia and
                  Derek Dunfield and
                  Carole{-}Jean Wu and
                  Pontus Stenetorp and
                  Nicola Cancedda and
                  Jakob Nicolaus Foerster and
                  Yoram Bachrach},
  title        = {{AI} Research Agents for Machine Learning: Search, Exploration, and
                  Generalization in MLE-bench},
  journal      = {CoRR},
  volume       = {abs/2507.02554},
  year         = {2025},
  url          = {https://doi.org/10.48550/arXiv.2507.02554},
  doi          = {10.48550/ARXIV.2507.02554},
  eprinttype    = {arXiv},
  eprint       = {2507.02554},
  bibsource    = {dblp computer science bibliography, https://dblp.org}
}

@article {Swanson2024.11.11.623004,
	author = {Swanson, Kyle and Wu, Wesley and Bulaong, Nash L. and Pak, John E. and Zou, James},
	title = {The Virtual Lab: AI Agents Design New SARS-CoV-2 Nanobodies with Experimental Validation},
	elocation-id = {2024.11.11.623004},
	year = {2024},
	doi = {10.1101/2024.11.11.623004},
	publisher = {Cold Spring Harbor Laboratory},
	URL = {https://www.biorxiv.org/content/early/2024/11/12/2024.11.11.623004},
	eprint = {https://www.biorxiv.org/content/early/2024/11/12/2024.11.11.623004.full.pdf},
	journal = {bioRxiv}
}

@article{DBLP:journals/corr/abs-2408-06292,
  author       = {Chris Lu and
                  Cong Lu and
                  Robert Tjarko Lange and
                  Jakob N. Foerster and
                  Jeff Clune and
                  David Ha},
  title        = {The {AI} Scientist: Towards Fully Automated Open-Ended Scientific
                  Discovery},
  journal      = {CoRR},
  volume       = {abs/2408.06292},
  year         = {2024},
  url          = {https://doi.org/10.48550/arXiv.2408.06292},
  doi          = {10.48550/ARXIV.2408.06292},
  eprinttype    = {arXiv},
  eprint       = {2408.06292},
  bibsource    = {dblp computer science bibliography, https://dblp.org}
}

@article{Jiang2025AIDEAE,
  title={AIDE: AI-Driven Exploration in the Space of Code},
  author={Zhengyao Jiang and Dominik Schmidt and Dhruv Srikanth and Dixing Xu and Ian Kaplan and Deniss Jacenko and Yuxiang Wu},
  journal={ArXiv},
  year={2025},
  volume={abs/2502.13138},
  url={https://api.semanticscholar.org/CorpusID:276421281}
}

@article{DBLP:journals/corr/abs-2505-09388,
  author       = {An Yang and
                  Anfeng Li and
                  Baosong Yang and
                  Beichen Zhang and
                  Binyuan Hui and
                  Bo Zheng and
                  Bowen Yu and
                  Chang Gao and
                  Chengen Huang and
                  Chenxu Lv and
                  Chujie Zheng and
                  Dayiheng Liu and
                  Fan Zhou and
                  Fei Huang and
                  Feng Hu and
                  Hao Ge and
                  Haoran Wei and
                  Huan Lin and
                  Jialong Tang and
                  Jian Yang and
                  Jianhong Tu and
                  Jianwei Zhang and
                  Jian Yang and
                  Jiaxi Yang and
                  Jingren Zhou and
                  Junyang Lin and
                  Kai Dang and
                  Keqin Bao and
                  Kexin Yang and
                  Le Yu and
                  Lianghao Deng and
                  Mei Li and
                  Mingfeng Xue and
                  Mingze Li and
                  Pei Zhang and
                  Peng Wang and
                  Qin Zhu and
                  Rui Men and
                  Ruize Gao and
                  Shixuan Liu and
                  Shuang Luo and
                  Tianhao Li and
                  Tianyi Tang and
                  Wenbiao Yin and
                  Xingzhang Ren and
                  Xinyu Wang and
                  Xinyu Zhang and
                  Xuancheng Ren and
                  Yang Fan and
                  Yang Su and
                  Yichang Zhang and
                  Yinger Zhang and
                  Yu Wan and
                  Yuqiong Liu and
                  Zekun Wang and
                  Zeyu Cui and
                  Zhenru Zhang and
                  Zhipeng Zhou and
                  Zihan Qiu},
  title        = {Qwen3 Technical Report},
  journal      = {CoRR},
  volume       = {abs/2505.09388},
  year         = {2025},
  url          = {https://doi.org/10.48550/arXiv.2505.09388},
  doi          = {10.48550/ARXIV.2505.09388},
  eprinttype    = {arXiv},
  eprint       = {2505.09388},
  bibsource    = {dblp computer science bibliography, https://dblp.org}
}

@article{DBLP:journals/corr/abs-2506-07976,
  author       = {Junhong Shen and
                  Hao Bai and
                  Lunjun Zhang and
                  Yifei Zhou and
                  Amrith Setlur and
                  Shengbang Tong and
                  Diego Caples and
                  Nan Jiang and
                  Tong Zhang and
                  Ameet Talwalkar and
                  Aviral Kumar},
  title        = {Thinking vs. Doing: Agents that Reason by Scaling Test-Time Interaction},
  journal      = {CoRR},
  volume       = {abs/2506.07976},
  year         = {2025},
  url          = {https://doi.org/10.48550/arXiv.2506.07976},
  doi          = {10.48550/ARXIV.2506.07976},
  eprinttype    = {arXiv},
  eprint       = {2506.07976},
  bibsource    = {dblp computer science bibliography, https://dblp.org}
}

@inproceedings{DBLP:conf/eurosys/ShengZYWZZPL025,
  author       = {Guangming Sheng and
                  Chi Zhang and
                  Zilingfeng Ye and
                  Xibin Wu and
                  Wang Zhang and
                  Ru Zhang and
                  Yanghua Peng and
                  Haibin Lin and
                  Chuan Wu},
  title        = {HybridFlow: {A} Flexible and Efficient {RLHF} Framework},
  booktitle    = {Proceedings of the Twentieth European Conference on Computer Systems,
                  EuroSys 2025, Rotterdam, The Netherlands, 30 March 2025 - 3 April
                  2025},
  pages        = {1279--1297},
  publisher    = {{ACM}},
  year         = {2025},
  url          = {https://doi.org/10.1145/3689031.3696075},
  doi          = {10.1145/3689031.3696075},
  bibsource    = {dblp computer science bibliography, https://dblp.org}
}

@inproceedings{DBLP:conf/naacl/AnugrahaWLIL25,
  author       = {David Anugraha and
                  Genta Indra Winata and
                  Chenyue Li and
                  Patrick Amadeus Irawan and
                  En{-}Shiun Annie Lee},
  editor       = {Luis Chiruzzo and
                  Alan Ritter and
                  Lu Wang},
  title        = {ProxyLM: Predicting Language Model Performance on Multilingual Tasks
                  via Proxy Models},
  booktitle    = {Findings of the Association for Computational Linguistics: {NAACL}
                  2025, Albuquerque, New Mexico, USA, April 29 - May 4, 2025},
  pages        = {1981--2011},
  publisher    = {Association for Computational Linguistics},
  year         = {2025},
  url          = {https://doi.org/10.18653/v1/2025.findings-naacl.106},
  doi          = {10.18653/V1/2025.FINDINGS-NAACL.106},
  bibsource    = {dblp computer science bibliography, https://dblp.org}
}

@misc{park2025lookleapestimatingllm,
      title={Look Before you Leap: Estimating LLM Benchmark Scores from Descriptions}, 
      author={Jungsoo Park and Ethan Mendes and Gabriel Stanovsky and Alan Ritter},
      year={2025},
      eprint={2509.20645},
      archivePrefix={arXiv},
      primaryClass={cs.CL},
      url={https://arxiv.org/abs/2509.20645}, 
}

@inproceedings{DBLP:conf/iclr/HuSWALWWC22,
  author       = {Edward J. Hu and
                  Yelong Shen and
                  Phillip Wallis and
                  Zeyuan Allen{-}Zhu and
                  Yuanzhi Li and
                  Shean Wang and
                  Lu Wang and
                  Weizhu Chen},
  title        = {LoRA: Low-Rank Adaptation of Large Language Models},
  booktitle    = {The Tenth International Conference on Learning Representations, {ICLR}
                  2022, Virtual Event, April 25-29, 2022},
  publisher    = {OpenReview.net},
  year         = {2022},
  url          = {https://openreview.net/forum?id=nZeVKeeFYf9},
  bibsource    = {dblp computer science bibliography, https://dblp.org}
}

@article{DBLP:journals/corr/abs-2408-03314,
  author       = {Charlie Snell and
                  Jaehoon Lee and
                  Kelvin Xu and
                  Aviral Kumar},
  title        = {Scaling {LLM} Test-Time Compute Optimally can be More Effective than
                  Scaling Model Parameters},
  journal      = {CoRR},
  volume       = {abs/2408.03314},
  year         = {2024},
  url          = {https://doi.org/10.48550/arXiv.2408.03314},
  doi          = {10.48550/ARXIV.2408.03314},
  eprinttype    = {arXiv},
  eprint       = {2408.03314},
  bibsource    = {dblp computer science bibliography, https://dblp.org}
}

@article{DBLP:journals/corr/abs-2506-12928,
  author       = {King Zhu and
                  Hanhao Li and
                  Siwei Wu and
                  Tianshun Xing and
                  Dehua Ma and
                  Xiangru Tang and
                  Minghao Liu and
                  Jian Yang and
                  Jiaheng Liu and
                  Yuchen Eleanor Jiang and
                  Changwang Zhang and
                  Chenghua Lin and
                  Jun Wang and
                  Ge Zhang and
                  Wangchunshu Zhou},
  title        = {Scaling Test-time Compute for {LLM} Agents},
  journal      = {CoRR},
  volume       = {abs/2506.12928},
  year         = {2025},
  url          = {https://doi.org/10.48550/arXiv.2506.12928},
  doi          = {10.48550/ARXIV.2506.12928},
  eprinttype    = {arXiv},
  eprint       = {2506.12928},
  bibsource    = {dblp computer science bibliography, https://dblp.org}
}

@article{DBLP:journals/corr/abs-2503-14476,
  author       = {Qiying Yu and
                  Zheng Zhang and
                  Ruofei Zhu and
                  Yufeng Yuan and
                  Xiaochen Zuo and
                  Yu Yue and
                  Tiantian Fan and
                  Gaohong Liu and
                  Lingjun Liu and
                  Xin Liu and
                  Haibin Lin and
                  Zhiqi Lin and
                  Bole Ma and
                  Guangming Sheng and
                  Yuxuan Tong and
                  Chi Zhang and
                  Mofan Zhang and
                  Wang Zhang and
                  Hang Zhu and
                  Jinhua Zhu and
                  Jiaze Chen and
                  Jiangjie Chen and
                  Chengyi Wang and
                  Hongli Yu and
                  Weinan Dai and
                  Yuxuan Song and
                  Xiangpeng Wei and
                  Hao Zhou and
                  Jingjing Liu and
                  Wei{-}Ying Ma and
                  Ya{-}Qin Zhang and
                  Lin Yan and
                  Mu Qiao and
                  Yonghui Wu and
                  Mingxuan Wang},
  title        = {{DAPO:} An Open-Source {LLM} Reinforcement Learning System at Scale},
  journal      = {CoRR},
  volume       = {abs/2503.14476},
  year         = {2025},
  url          = {https://doi.org/10.48550/arXiv.2503.14476},
  doi          = {10.48550/ARXIV.2503.14476},
  eprinttype    = {arXiv},
  eprint       = {2503.14476},
  bibsource    = {dblp computer science bibliography, https://dblp.org}
}

@article{DBLP:journals/corr/abs-2402-03300,
  author       = {Zhihong Shao and
                  Peiyi Wang and
                  Qihao Zhu and
                  Runxin Xu and
                  Junxiao Song and
                  Mingchuan Zhang and
                  Y. K. Li and
                  Y. Wu and
                  Daya Guo},
  title        = {DeepSeekMath: Pushing the Limits of Mathematical Reasoning in Open
                  Language Models},
  journal      = {CoRR},
  volume       = {abs/2402.03300},
  year         = {2024},
  url          = {https://doi.org/10.48550/arXiv.2402.03300},
  doi          = {10.48550/ARXIV.2402.03300},
  eprinttype    = {arXiv},
  eprint       = {2402.03300},
  bibsource    = {dblp computer science bibliography, https://dblp.org}
}

@inproceedings{DBLP:conf/icml/BurnsIKBGACEJLS24,
  author       = {Collin Burns and
                  Pavel Izmailov and
                  Jan Hendrik Kirchner and
                  Bowen Baker and
                  Leo Gao and
                  Leopold Aschenbrenner and
                  Yining Chen and
                  Adrien Ecoffet and
                  Manas Joglekar and
                  Jan Leike and
                  Ilya Sutskever and
                  Jeffrey Wu},
  title        = {Weak-to-Strong Generalization: Eliciting Strong Capabilities With
                  Weak Supervision},
  booktitle    = {Forty-first International Conference on Machine Learning, {ICML} 2024,
                  Vienna, Austria, July 21-27, 2024},
  publisher    = {OpenReview.net},
  year         = {2024},
  url          = {https://openreview.net/forum?id=ghNRg2mEgN},
  bibsource    = {dblp computer science bibliography, https://dblp.org}
}

@article{DBLP:journals/corr/abs-2505-23723,
  author       = {Zexi Liu and
                  Jingyi Chai and
                  Xinyu Zhu and
                  Shuo Tang and
                  Rui Ye and
                  Bolun Zhang and
                  Lei Bai and
                  Siheng Chen},
  title        = {ML-Agent: Reinforcing {LLM} Agents for Autonomous Machine Learning
                  Engineering},
  journal      = {CoRR},
  volume       = {abs/2505.23723},
  year         = {2025},
  url          = {https://doi.org/10.48550/arXiv.2505.23723},
  doi          = {10.48550/ARXIV.2505.23723},
  eprinttype    = {arXiv},
  eprint       = {2505.23723},
  bibsource    = {dblp computer science bibliography, https://dblp.org}
}

@article{DBLP:journals/corr/abs-2505-07782,
  author       = {Rushi Qiang and
                  Yuchen Zhuang and
                  Yinghao Li and
                  Dingu Sagar V. K and
                  Rongzhi Zhang and
                  Changhao Li and
                  Ian Shu{-}Hei Wong and
                  Sherry Yang and
                  Percy Liang and
                  Chao Zhang and
                  Bo Dai},
  title        = {MLE-Dojo: Interactive Environments for Empowering {LLM} Agents in
                  Machine Learning Engineering},
  journal      = {CoRR},
  volume       = {abs/2505.07782},
  year         = {2025},
  url          = {https://doi.org/10.48550/arXiv.2505.07782},
  doi          = {10.48550/ARXIV.2505.07782},
  eprinttype    = {arXiv},
  eprint       = {2505.07782},
  bibsource    = {dblp computer science bibliography, https://dblp.org}
}

@inproceedings{DBLP:conf/iclr/ChanCJASMSLMPMW25,
  author       = {Jun Shern Chan and
                  Neil Chowdhury and
                  Oliver Jaffe and
                  James Aung and
                  Dane Sherburn and
                  Evan Mays and
                  Giulio Starace and
                  Kevin Liu and
                  Leon Maksin and
                  Tejal Patwardhan and
                  Aleksander Madry and
                  Lilian Weng},
  title        = {MLE-bench: Evaluating Machine Learning Agents on Machine Learning
                  Engineering},
  booktitle    = {The Thirteenth International Conference on Learning Representations,
                  {ICLR} 2025, Singapore, April 24-28, 2025},
  publisher    = {OpenReview.net},
  year         = {2025},
  url          = {https://openreview.net/forum?id=6s5uXNWGIh},
  bibsource    = {dblp computer science bibliography, https://dblp.org}
}

@inproceedings{Huang2023MLAgentBenchEL,
  author       = {Qian Huang and
                  Jian Vora and
                  Percy Liang and
                  Jure Leskovec},
  title        = {MLAgentBench: Evaluating Language Agents on Machine Learning Experimentation},
  booktitle    = {Forty-first International Conference on Machine Learning, {ICML} 2024,
                  Vienna, Austria, July 21-27, 2024},
  publisher    = {OpenReview.net},
  year         = {2024},
  url          = {https://openreview.net/forum?id=1Fs1LvjYQW},
  bibsource    = {dblp computer science bibliography, https://dblp.org}
}

@inproceedings{DBLP:conf/nips/YangJWLYNP24,
  author       = {John Yang and
                  Carlos E. Jimenez and
                  Alexander Wettig and
                  Kilian Lieret and
                  Shunyu Yao and
                  Karthik Narasimhan and
                  Ofir Press},
  editor       = {Amir Globersons and
                  Lester Mackey and
                  Danielle Belgrave and
                  Angela Fan and
                  Ulrich Paquet and
                  Jakub M. Tomczak and
                  Cheng Zhang},
  title        = {SWE-agent: Agent-Computer Interfaces Enable Automated Software Engineering},
  booktitle    = {Advances in Neural Information Processing Systems 38: Annual Conference
                  on Neural Information Processing Systems 2024, NeurIPS 2024, Vancouver,
                  BC, Canada, December 10 - 15, 2024},
  year         = {2024},
  url          = {http://papers.nips.cc/paper\_files/paper/2024/hash/5a7c947568c1b1328ccc5230172e1e7c-Abstract-Conference.html},
  bibsource    = {dblp computer science bibliography, https://dblp.org}
}

@inproceedings{DBLP:conf/iclr/JimenezYWYPPN24,
  author       = {Carlos E. Jimenez and
                  John Yang and
                  Alexander Wettig and
                  Shunyu Yao and
                  Kexin Pei and
                  Ofir Press and
                  Karthik R. Narasimhan},
  title        = {SWE-bench: Can Language Models Resolve Real-world Github Issues?},
  booktitle    = {The Twelfth International Conference on Learning Representations,
                  {ICLR} 2024, Vienna, Austria, May 7-11, 2024},
  publisher    = {OpenReview.net},
  year         = {2024},
  url          = {https://openreview.net/forum?id=VTF8yNQM66},
  bibsource    = {dblp computer science bibliography, https://dblp.org}
}

@inproceedings{opendevin,
  author       = {Xingyao Wang and
                  Boxuan Li and
                  Yufan Song and
                  Frank F. Xu and
                  Xiangru Tang and
                  Mingchen Zhuge and
                  Jiayi Pan and
                  Yueqi Song and
                  Bowen Li and
                  Jaskirat Singh and
                  Hoang H. Tran and
                  Fuqiang Li and
                  Ren Ma and
                  Mingzhang Zheng and
                  Bill Qian and
                  Yanjun Shao and
                  Niklas Muennighoff and
                  Yizhe Zhang and
                  Binyuan Hui and
                  Junyang Lin and
                  et al.},
  title        = {OpenHands: An Open Platform for {AI} Software Developers as Generalist
                  Agents},
  booktitle    = {The Thirteenth International Conference on Learning Representations,
                  {ICLR} 2025, Singapore, April 24-28, 2025},
  publisher    = {OpenReview.net},
  year         = {2025},
  url          = {https://openreview.net/forum?id=OJd3ayDDoF},
  bibsource    = {dblp computer science bibliography, https://dblp.org}
}
\appendix

\clearpage

\section{Ideation RL Training Formulation}
\subsection{Step-Level RL Objective}
\label{app:rl_obj}
We train the \textsc{Ideator}'s policy, $\pi_{\theta}(a|s)$, which is parameterized by a small language model with parameters $\theta$. Following recent work, we adopt a step-level RL paradigm for efficient training. The objective is to find the optimal parameters $\theta^*$ that maximize the expected reward from a single-step rollout:
$$
\theta^* = \arg\max_{\theta} \mathbb{E}_{s_t \sim d^{\pi_e}, \alpha \sim \pi_{\theta}(\cdot|s_t)} [R_t(\alpha)]
$$
where $d^{\pi_e}$ represents the state distribution of the expert coding agent. This stepwise objective is efficient as it allows us to reuse an offline buffer of trajectories and avoids costly multi-step rollouts.

\subsection{GRPO Loss Function}
\label{app:grpo_loss}

We formulate our GRPO training loss function~\cite{DBLP:journals/corr/abs-2402-03300} as follows, while removing the KL divergence term from the loss following recent practice~\cite{DBLP:journals/corr/abs-2503-14476}.

\begin{align*}
& \mathcal{J}_{\text{GRPO}}(\theta) = \mathbb{E}_{\substack{s_t \sim d^{e,c} \\ \{\alpha_i\}_{i=1}^G \sim \pi_{\theta_{\text{old}}}(\cdot|s_t)}} \Bigg[ \frac{1}{G} \sum_{i=1}^{G} \frac{1}{|\alpha_i|} \sum_{j=1}^{|\alpha_i|} \\
& \min\left(r_{i,j}(\theta) A_{i,j}, \text{clip}(r_{i,j}(\theta), 1-\epsilon, 1+\epsilon)A_{i,j}\right) \Bigg],
\end{align*}
\begin{gather*}
\text{with } r_{i,j}(\theta) = \frac{\pi_{\theta}(\alpha_{i,j}|s_t, \alpha_{i, <j})}{\pi_{\theta_{\text{old}}}(\alpha_{i,j}|s_t, \alpha_{i, <j})} \\
\text{and } A_{i,j} = \frac{R_t(\alpha_i) - \text{mean}(\{R_t(\alpha_i)\}_{i=1}^G)}{\text{std}(\{R_t(\alpha_i)\}_{i=1}^G)}.
\end{gather*}

\section{Training Details}
\subsection{Training Task Selection}
\label{app:training_task_selection}
For training, we selected 10 tasks spanning diverse modalities—text, image, audio, video, and tabular—whose solutions can typically be executed in under 20 minutes. They are:
\begin{itemize}[noitemsep, topsep=2pt]
    \item  osic-pulmonary-fibrosis-progression
    \item  multi-modal-gesture-recognition
    \item  chaii-hindi-and-tamil-question-answering
    \item  mlsp-2013-birds
    \item  google-quest-challenge
    \item  tgs-salt-identification-challenge
    \item  tweet-sentiment-extraction
    \item  spaceship-titanic
    \item  jigsaw-unintended-bias-in-toxicity-classification
    \item  AI4Code
\end{itemize}

\subsection{Training Hyperparameters}
\label{app:training_hyperparam}
We use VeRL~\cite{DBLP:conf/eurosys/ShengZYWZZPL025} as our RL training framework. We train the Qwen3-8B model with LoRA~\cite{DBLP:conf/iclr/HuSWALWWC22} using a rank of 32 for one epoch on a single node (8x A100 40GB GPUs) with a batch size of 16 and a rollout size ($G$) of 8 candidate ideas per prompt. We save checkpoints every 10 steps, selecting the best one based on validation reward. The reward calculation is distributed across 16 nodes, each with 8 A10G (24GB) GPUs, where each GPU is dedicated to running the solution for a single candidate idea. The total training time is approximately 52 hours.

\begin{figure}
    \centering 
    \includegraphics[width=\linewidth]{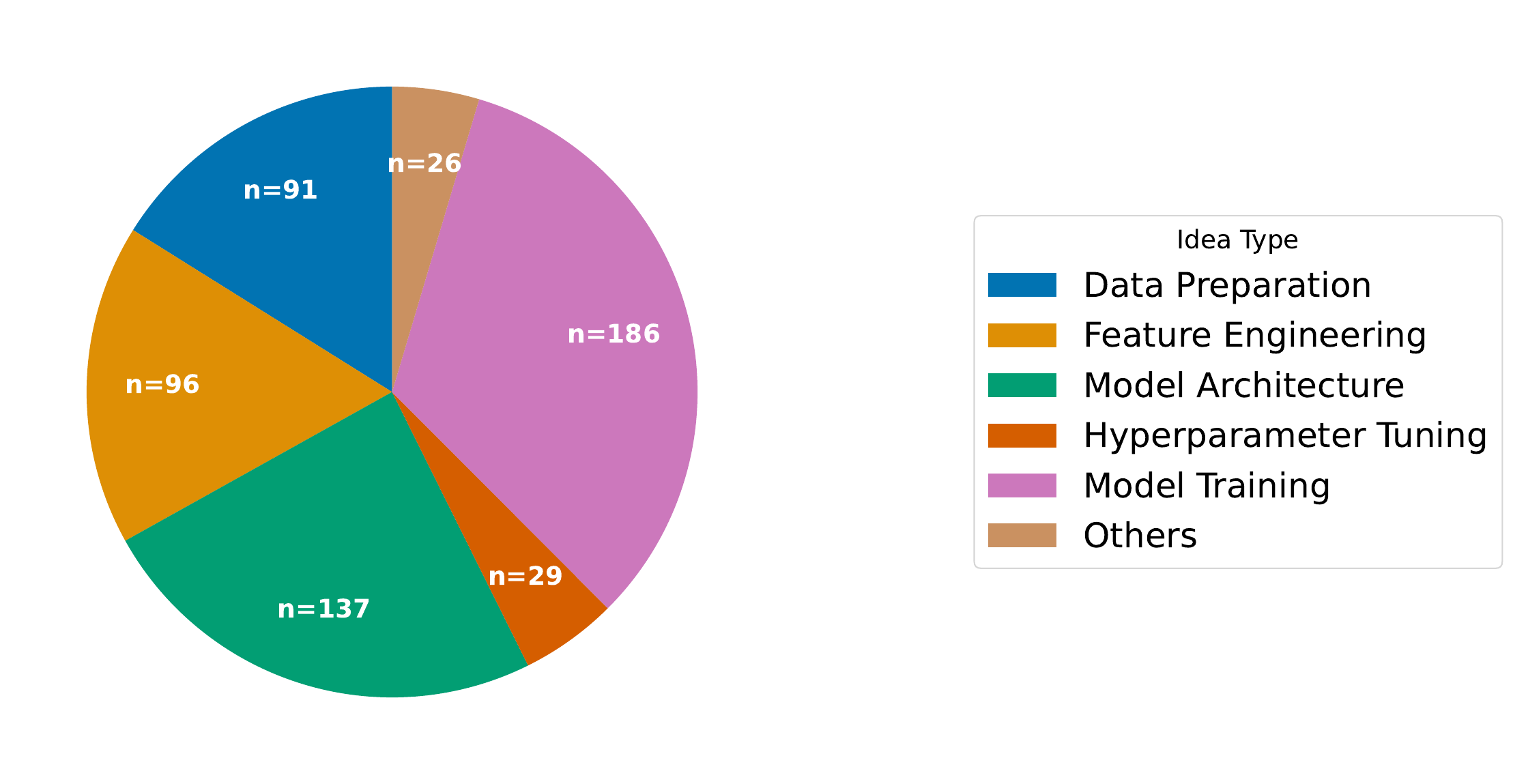}
    \caption{Distribution of idea types, aggregated across Qwen3-8B, Qwen3-8B-RL and Claude Sonnet 3.5 as \textsc{Ideators}, paired with Claude Sonnet 3.5 as the implementer.} 
    \label{fig:idea_type_distribution} 
\end{figure}

\begin{figure}
    \centering 
    \includegraphics[width=\linewidth]{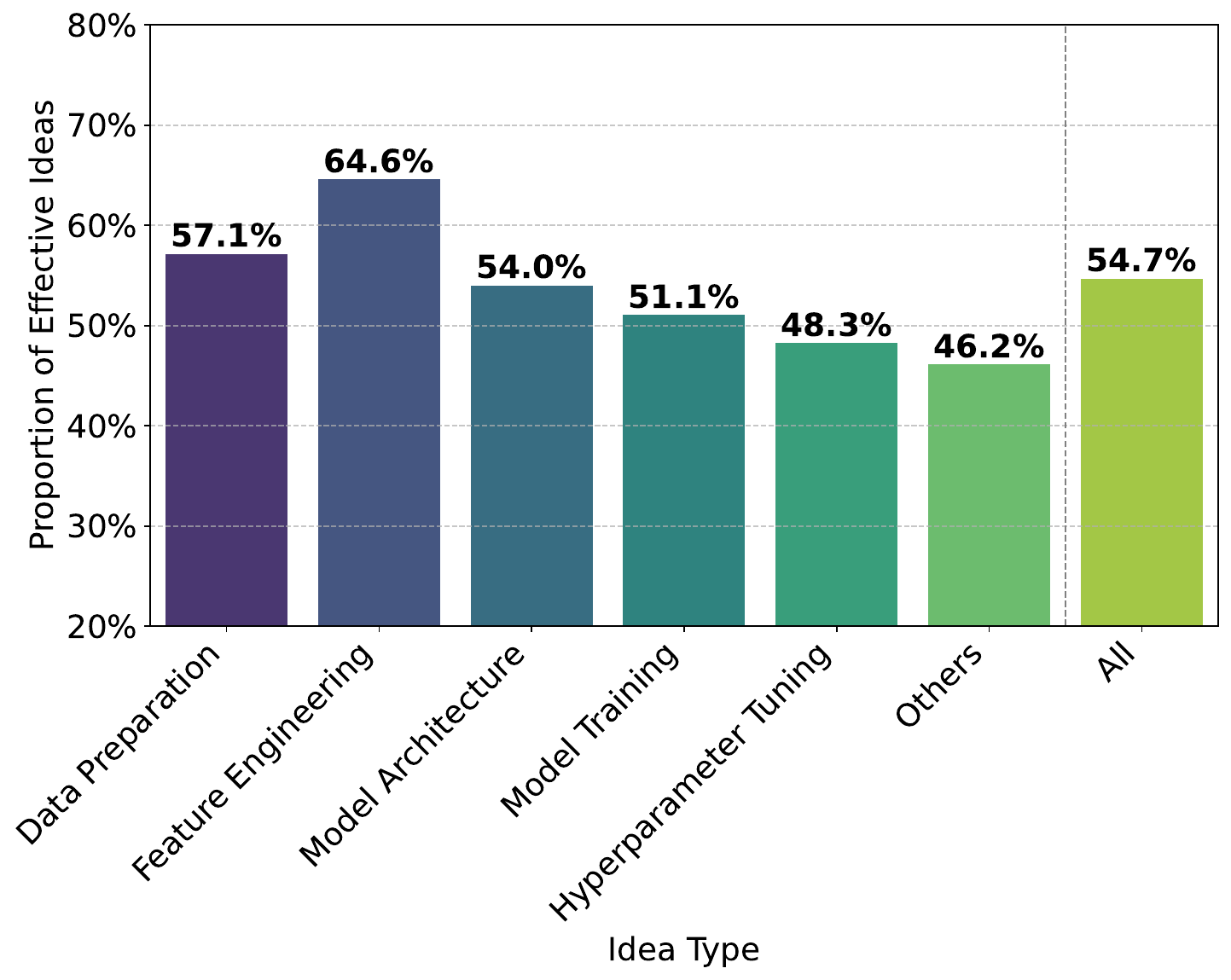}
    \caption{Proportion of effective ideas in each idea type, aggregated across Qwen3-8B, Qwen3-8B-RL and Claude Sonnet 3.5 as \textsc{Ideators}, paired with Claude Sonnet 3.5 as the Implementer. We define an effective idea as the performance of the refined ML solution according to the idea is better than before.} 
    \label{fig:idea_effectiveness} 
\end{figure}

\section{Distribution and Effectiveness of Idea Types}
\label{app:idea_type_analysis}
Figure~\ref{fig:idea_type_distribution} shows the overall distribution of ideas. The most frequent suggestions concerned \textbf{Model Training}, followed by \textbf{Model Architecture} and \textbf{Feature Engineering}. However, frequency alone does not imply effectiveness. We define an idea as \textit{effective} if it improved the ML solution’s performance. As shown in Figure~\ref{fig:idea_effectiveness}, \textbf{Feature Engineering} had the highest success rate (64.6\%), followed by \textbf{Data Preparation} (57.1\%). In contrast, \textbf{Model Training}, despite being the most frequent category, was only the fourth most effective (51.1\%), while \textbf{Hyperparameter Tuning} was least effective (48.3\%). This highlights a central insight: \textit{the most impactful interventions often concern data and features rather than the model itself}.

\section{Prompt for the \textsc{Ideator}}
\label{app:prompt_ideator}

The specific instructions provided to the \textsc{Ideator} agent are detailed below. This prompt is designed to guide the model in generating high-quality, actionable ideas, while also specifying the required output format.

\begin{tcolorbox}[
    breakable,
    colback=black!5!white,         
    colframe=black!75!white,       
    title=Idea Generation Prompt, 
    fonttitle=\bfseries,           
    boxrule=1pt,                   
    arc=2mm,                       
]
\small
You are a machine learning expert. Another AI agent is struggling to improve the performance of its machine learning solution. Your task is to analyze the agent's progress and provide the most effective algorithmic idea that can significantly improve the performance.
\\\\
You will be provided with the agent's history, including its previous attempts and the full trajectory of its actions.
\\\\
\{query\}
\\\\
TRAJECTORY:
\\
\{step\_trace\}
\\\\
\#\#\# Instruction

Evaluate the current trajectory, pick the **single highest-impact next action**, and then output **exactly three items** in this format:
\\

ANALYSIS\_ON\_CURRENT\_PROGRESS:\\
<Briefly state whether to keep refining the present approach or revert to a prior solution and pursue a new path.>
\\\\
ACTION:\\
<One imperative command or code block the agent must execute next.>
\\\\
RATIONALE:\\
<Concise justification for why this action is optimal right now.>
\\

Do **not** list alternatives, background, or extra commentary. Output nothing beyond those three items.

\end{tcolorbox}

\section{Prompt for Idea Type Classification}
\label{app:prompt_classification}
We utilize the following prompt to instruct Claude Sonnet 4 in categorizing the generated ideas. The prompt includes detailed definitions for each idea type and provides the model with rubrics to distinguish between categories.

\begin{tcolorbox}[
    breakable,
    colback=black!5!white,         
    colframe=black!75!white,       
    title=Idea Type Classification Prompt, 
    fonttitle=\bfseries,           
    boxrule=1pt,                   
    arc=2mm,                       
]
\small
You are a machine learning expert. Another AI agent is working on a task and has encountered a problem. Your goal is to analyze a proposed idea for improving a machine learning model and classify it into one of the specific categories defined below.
\\\\
\{trace\}
\\\\
\#\#\# Proposed Idea
\\\\
\{idea\}
\\\\

Your Task:\\
- Review the trajectory to understand the agent's attempts so far to solve the task.\\
- Consider the problem description and the proposed idea.\\
- Determine which category (from a predefined list below) best describes the core focus of the idea.\\
- Provide a brief rationale for your classification.\\
\\\\
* Data Preparation: Ideas for handling missing values, correcting errors, removing outliers, or augmenting existing data.\\
* Feature Engineering: Ideas for creating new predictive features, transforming existing features (e.g., scaling, encoding), or selecting the most important ones.\\
* Model Architecture: Ideas for switching to a completely different type of model (e.g., from XGBoost to a Neural Network) or changing the fundamental structure of the current model (e.g., adding/removing layers).\\
* Hyperparameter Tuning: Ideas for systematically searching for the best model settings (e.g., learning rate, tree depth) to improve performance.\\
* Model Training: Ideas for changing the training process (e.g., using a different loss function, implementing cross-validation) or using new metrics to evaluate the model.\\
* Others: Use this category only if the idea clearly does not fit into any of the categories above.
\\\\
Output Format: output **exactly two items** in this format:\\
\\\\
TYPE: <one of "Data Preparation", "Feature Engineering", "Model Architecture", "Hyperparameter Tuning", "Model Training", or "Others">\\
\\
RATIONALE: <briefly explain why the idea belongs to the type above>\\
\\\\
Output nothing beyond those two items.
\end{tcolorbox}

\section{Case Studies on Ideation}
\label{app:case_study}

\subsection{An Effective Feature Engineering Idea}

 In the \textit{random-acts-of-pizza} task where numerical features showed very low importance ($\sim$1–1.5\%), the Ideator suggested creating interaction-based derived features such as \texttt{karma\_ratio}, \texttt{activity\_ratio}, \texttt{engagement\_score}, \texttt{text\_length}, and \texttt{time\_of\_day}. These features added contextual structure that the raw values lacked, and the Implementer’s accuracy improved from 0.60 $\to$ 0.64.

\begin{tcolorbox}[breakable, colback=implementerBack, colframe=implementerFrame, title=\textbf{Implementer Query}]
\textbf{PROBLEM\_STATEMENT:}\\
Initial model shows relatively low feature importance for numerical features, need to improve model performance

\textbf{ATTEMPTS\_SO\_FAR:}
\begin{itemize}[noitemsep,topsep=0pt]
    \item Created baseline with TF-IDF text features and basic numerical features
    \item Used RandomForestClassifier with default parameters
    \item Feature importance shows numerical features each contributing only $\sim$1-1.5\%
\end{itemize}

\textbf{GOAL:}\\
Identify ways to improve model performance
\end{tcolorbox}

\begin{tcolorbox}[breakable, colback=ideatorBack, colframe=ideatorFrame, title=\textbf{Ideator Response}]
\textbf{ANALYSIS\_ON\_CURRENT\_PROGRESS:}\\
Keep refining the present approach, as the foundation with text and metadata features is sound but needs optimization.

\textbf{ACTION:}
\begin{lstlisting}
# Create derived features that capture interaction effects:
karma_ratio = upvotes_minus_downvotes / upvotes_plus_downvotes
activity_ratio = comments_in_raop / total_comments
engagement_score = number_of_comments * karma_ratio
text_length = len(request_text)
time_of_day = extract hour from unix_timestamp
\end{lstlisting}

\textbf{RATIONALE:}\\
Raw numerical features show low importance because they lack context - derived features that capture relationships between metrics will better represent user credibility and request quality patterns.
\end{tcolorbox}

\subsection{An Ineffective Hyperparameter Tuning Idea}

In the \textit{whale-categorization-playground} task, the Implementer had already reached 72.31\% MAP@5 with a ViT model and combined Focal+ArcFace loss. The Ideator proposed further tuning ArcFace parameters (scale, margin, alpha, gamma), but this yielded no benefit and even reduced MAP@5 (0.30 $\to$ 0.21). This reflects a common pattern: when the model is already near its capacity and only minor hyperparameters remain, ideation provides limited gains.

\begin{tcolorbox}[breakable, colback=implementerBack, colframe=implementerFrame, title=\textbf{Implementer Query}]
\textbf{PROBLEM\_STATEMENT:}\\
Need to further improve MAP@5 score with current ViT + Combined Loss approach that achieved 72.31\% accuracy

\textbf{ATTEMPTS\_SO\_FAR:}
\begin{itemize}[noitemsep,topsep=0pt]
    \item Used EfficientNet-B0/B3 with augmentations ($\sim$10\% accuracy)
    \item Implemented Focal Loss and ArcFace separately
    \item Combined approach with ViT + Custom Loss ($\sim$17\% accuracy initially)
    \item Adjusted learning rates and increased epochs (72.31\% accuracy)
    \item Generated valid submission file
\end{itemize}

\textbf{GOAL:}\\
Further optimize model performance by either enhancing the current approach or exploring complementary techniques while maintaining the valid submission format
\end{tcolorbox}

\begin{tcolorbox}[breakable, colback=ideatorBack, colframe=ideatorFrame, title=\textbf{Ideator Response}]
\textbf{ANALYSIS\_ON\_CURRENT\_PROGRESS:}\\
The current ViT model with combined Focal and ArcFace loss has achieved the same accuracy as the initial ViT approach, suggesting the loss function may not be optimally balanced. Further tuning of the loss parameters or exploring advanced regularization techniques could yield improvements.

\textbf{ACTION:}
\begin{lstlisting}
criterion = CombinedLoss(num_classes=len(train_dataset.id_to_idx), embedding_size=768, scale=60.0, margin=0.3, alpha=0.5, gamma=4)
\end{lstlisting}

\textbf{RATIONALE:}\\
Increasing the ArcFace scale factor to 60.0 improves the margin between classes, while adjusting alpha and gamma to 0.5 and 4 respectively enhances focus on hard examples. This directly addresses class imbalance and improves feature discrimination, which is critical for the task.
\end{tcolorbox}

\end{document}